\documentclass[11pt]{article}

\usepackage[final]{acl}

\usepackage{times}
\usepackage{latexsym}

\usepackage[T1]{fontenc}

\usepackage[utf8]{inputenc}

\usepackage{microtype}

\usepackage{inconsolata}

\usepackage{graphicx}

\usepackage{booktabs}
\usepackage{makecell} 
\usepackage{multirow} 
\usepackage{amsmath,amssymb}
\usepackage{subcaption} 
\usepackage{csquotes} 
\usepackage{adjustbox}
\usepackage{listings}
\usepackage{xurl}

\title{Generative Floor Plan Design with LLMs via Reinforcement Learning with Verifiable Rewards}

\author{
 \textbf{Luis Lara\textsuperscript{1}},
 \textbf{Aristides Milios\textsuperscript{1,2}},
 \textbf{Zhi Hao Luo\textsuperscript{1,3}},
 \textbf{Aditya Sharma\textsuperscript{1,3}},
\\
 \textbf{Ge Ya Luo\textsuperscript{1,2}},
 \textbf{Christopher Beckham\textsuperscript{1}},
 \textbf{Florian Golemo\textsuperscript{1}},
 \textbf{Christopher Pal\textsuperscript{1,2,3,4}}
\\
 \textsuperscript{1}Mila – Quebec AI Institute,
 \textsuperscript{2}Université de Montréal,
 \\
\textsuperscript{3}Polytechnique Montréal,
 \textsuperscript{4}Canada CIFAR AI Chair
\\
 \small{
   \textbf{Correspondence:} \href{mailto:luis.lara@mila.quebec}{luis.lara@mila.quebec}
 }
}

\begin{document}
\maketitle
\begin{abstract}
An AI system for professional floor plan design must precisely control room dimensions and areas while respecting the desired connectivity between rooms and maintaining functional and aesthetic quality.
Existing generative approaches focus primarily on respecting the requested connectivity between rooms, but do not support generating floor plans that respect numerical constraints. 
We introduce a text‑based floor plan generation approach that fine-tunes a large language model (LLM) on real plans and then applies reinforcement learning with verifiable rewards (RLVR) to improve adherence to topological and numerical constraints while discouraging invalid or overlapping outputs.
Furthermore, we design a set of constraint adherence metrics to systematically measure how generated floor plans align with user-defined constraints.
Our model generates floor plans that satisfy user-defined connectivity and numerical constraints and outperforms existing methods on Realism, Compatibility, and Diversity metrics. Across all tasks, our approach achieves at least a 94\% relative reduction in Compatibility compared with existing methods.\footnote{Project code is available at \url{https://github.com/ludolara/floor-plan-rlvr}}
Our results demonstrate that LLMs can effectively handle constraints in this setting, suggesting broader applications for text-based generative modeling.
\end{abstract}

\section{Introduction}

\begin{figure*}[t]
    \centering
    \includegraphics[width=0.98\textwidth]{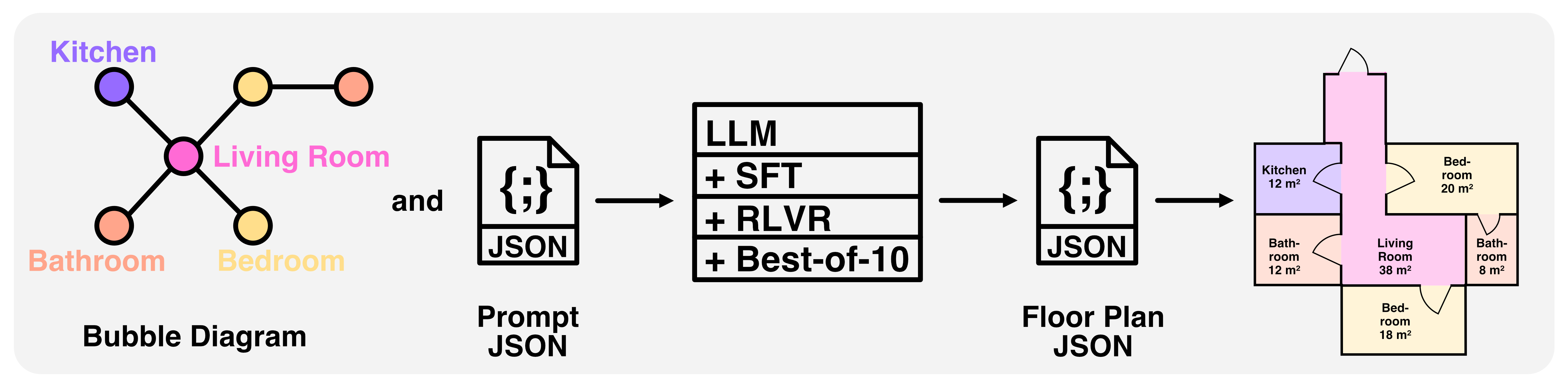}
    \caption{\textbf{Inference Process Overview.} Given a bubble diagram (input connectivity graph) and a JSON specification of design requirements (e.g., desired room sizes), our model generates a complete floor plan in JSON format.}
    \label{fig:training-overview}
\end{figure*}

Generative models have emerged as powerful tools for accelerating design across different sectors. However, their widespread adoption depends on their ability to strike a balance between flexibility and precise control. This challenge is particularly pronounced in floor plan generation, where users often need to specify strict constraints, including exact room sizing and connectivity. To address this, we propose a model that handles user-defined constraints, and we introduce evaluation metrics that measure how well generated floor plans satisfy those constraints.

Most existing generative modeling techniques for floor plan design rely on top-down 2D renderings and use input graphs, referred to as bubble diagrams, to represent the spatial connectivity between rooms.
However, such approaches have inherent drawbacks. 
Vision-based generative models, for instance, produce rasterized image-based outputs, making it challenging to access or modify them directly.
Additionally, many models focus exclusively on adjacency constraints, disregarding room geometry entirely, thereby reducing the degree of control users can exert over the generated layouts.

To overcome these limitations, we adopt a JSON-based representation for generative floor plan modeling that encodes room layouts as polygonal structures.
Our method builds on prior vectorized floor plan approaches, such as \textit{House-GAN}, \textit{House-GAN++} \citep{nauata2020house, nauata2021house}, and \textit{HouseDiffusion} \citep{shabani2023housediffusion}, and leverages data from the \textit{RPLAN} dataset \citep{rplan}.
This structured format facilitates greater control over spatial parameters, including room size and connectivity.
To validate this approach, we fine-tune a large language model (LLM) to generate JSON-encoded floor plans with spatial constraints specified in the prompt. See Figure \ref{fig:training-overview} for an overview of the inference process.

Through our experiments, we validate the effectiveness of our approach in maintaining high compliance with the specified constraints, as measured by our proposed constraint-adherence metrics.
In the most complex setting, the eight-room task, our approach with best-of-10 sampling reduces Compatibility by 94\% relative to \textit{HouseDiffusion} (Table~\ref{tab:gen_performance}). Compatibility is a standard metric of how well the generated floor plan matches the desired room connectivity, where lower is better.

Our key contributions are as follows:
\begin{enumerate}
    \item We fine-tune an LLM in two stages, first via supervised learning and then via reinforcement learning, to transform constraint inputs into valid structured floor plans, demonstrating the feasibility and advantages of a structured-data-to-structured-data generative paradigm.
    \item We propose new constraint adherence metrics that systematically evaluate the extent to which generated floor plans align with user-defined constraints, filling a critical evaluation gap in this domain.
\end{enumerate}

\section{Related Work}

\textbf{Generative Models.} \textit{House-GAN} and \textit{House-GAN++} \citep{nauata2020house,nauata2021house} are a family of GAN-based methods that learn to generate floor plan images using convolutional graph-based networks. 
 \textit{House-GAN} starts with a noise vector for every existing room in the connectivity graph and makes use of a convolutional message passing network (Conv-MPN) to update node features while preserving spatial relationships. 
Notably, however, these models only condition on a bubble diagram and cannot impose other forms of constraints. 
The outputs from these methods are also scale-invariant (i.e., not metric) and thus cannot be used directly in any downstream task. 
\textit{FloorplanGAN} \citep{FPGAN} proposes a self-attention-based GAN that takes as input room centers, desired areas (encoded as relative proportions), as well as room type for each room. 
However, the outputs from the GAN do not always adhere to the original constraints. 
While the use of a differentiable rasterizer to compute losses in pixel space opens up possibilities, the core method does not appear to support a partial specification of constraints or polygons. 
\textit{HouseDiffusion} \citep{shabani2023housediffusion} is a diffusion-based method that creates a 1-D polygonal loop for each room in the bubble diagram and, through the diffusion process, iteratively improves their shape and position. 
While it also utilizes a transformer architecture to attend to the input graph, the floor plan generation process is done through diffusion. 
This method also only allows the input of a bubble diagram without the additional ability to condition on numerical constraints and therefore suffers from the same shortcomings as the \textit{House-GAN} models. 

\textbf{Large Language Models.} \textit{ArchiText} \citep{galanos2023architext} also leverages LLMs to generate floor plans, but its prompts appear to be limited to natural-language descriptions rather than explicit geometry.
In contrast to these approaches, our method allows users to specify detailed room constraints (e.g., room areas) and outputs room polygons in metric units that can directly be used in CAD drawing software.
Providing both of these is crucial to the novelty and architectural usability of our method.

In \textit{Tell2Design} \cite{tell2design}, likely the most similar existing work, the authors create a new dataset for floor plan generation based on natural-language descriptions, based on \textit{RPLAN}. They incorporate spatial (sizing) and relational positioning constraints in their natural-language prompts, and train a T5 sequence-to-sequence model on their dataset. Unlike our work, the authors render the floor plans as images and use image-based metrics for evaluation, which functionally ignore structural errors such as overlaps. Additionally, the evaluation of constraint adherence is somewhat limited, with human evaluation of only 100 test samples along 4 different axes (room type, room location, room size, and room relationships). In their evaluation, they show that their trained T5 model struggles particularly with spatial relationships. In this work, we propose a robust set of complete metrics for constraint satisfaction that do not rely on rasterization, but rather measure the desired properties directly from the sequence output.

\textbf{3D Scene Generation.} Recently, full 3D scene generation methods have  shown impressive results.
\textit{AnyHome} \citep{wen2023anyhome} and \textit{Holodeck} \citep{yang2024holodeck} can generate floor plans, windows, doors, furniture, and meaningful placement in 3D, all from a natural-language prompt (e.g., "a 1b1b apartment of a researcher who has a cat").
In our method, we focus only on the floor plan aspect, but we allow for the specification of room dimensions and areas, as well as the total floor plan area, which neither method supports.

\textbf{Constraint Satisfaction and Symbolic Methods.} Earlier work formulates floor plan generation as a constraint satisfaction problem in which room geometries are represented by variables such as position and size, and layouts are generated by enforcing predefined spatial and dimensional rules \citep{medjdoub2000separating, li2000constraint, upasani2020automated}. Related work by \citet{lopes2010constrained} proposes a procedural approach that utilizes relative area targets rather than explicit dimensional constraints, and organizes the layout through a hierarchical system of zones, such as public and private areas, which restricts direct topological relations to rooms within the same zone. These methods rely on hand-crafted rules and manually defined structures rather than learned priors from data. In contrast, our method learns plausible layouts directly from real floor plans while still conditioning on explicit structured constraints.

\textbf{Datasets.} One of the most commonly used datasets for floor plan generation is \textit{RPLAN} \citep{rplan}. 
We use the dataset in this work, and we describe it in Section \ref{sec:data}. 
A limitation of \textit{RPLAN} for our setting is that it stores floor plans as images rather than in a vector format. As a result, it must first be converted into an intermediate structured representation.

\section{Representation Format}

Our task requires the model to satisfy explicit numerical and topological constraints and produce artifacts that can be automatically validated and consumed by downstream geometry and CAD tooling. Prior work shows that imposing structure in the \emph{conditioning signal} can improve reliability: standardized marker templates provide an explicit control interface that improves generation quality and instruction adherence \citep{dsouza-etal-2025-treasure}. Complementarily, work in semantic parsing and code generation shows that unconstrained free-form generation is prone to syntactically invalid or non-executable outputs, while enforcing structural constraints during generation improves validity and accuracy \citep{yin-neubig-2018-tranx,scholak-etal-2021-picard,raspanti-etal-2025-grammar}. Collectively, these results substantiate the use of explicit, structured representations for both the conditioning and output sides.

Accordingly, we choose JSON over alternative input and output formats such as natural-language specifications. JSON (1) enables unambiguous parsing of numerical constraints, reducing ambiguity in measurements and spatial relationships, (2) enforces a consistent structure across training examples, reducing the burden on the model to infer implicit fields, (3) naturally captures nested relationships in floor plan specifications via its hierarchical organization, and (4) supports direct integration with CAD tools and architectural software that already rely on structured representations. See Appendix~\ref{sec:json_schemas} for the input and output schemas.

\subsection{Dataset}
\label{sec:data}
\textbf{RPLAN} \citep{rplan} contains 80,788 real floor plans from Asia stored as $256\times256\times4$ images that encode boundaries and room labels. We convert each plan to a JSON layout using the \textit{House-GAN++} data reader\footnote{\url{https://github.com/sepidsh/Housegan-data-reader}} and a custom converter that reconstructs room polygons, assigns semantic labels, scales coordinates to meters, computes geometric attributes, and derives a bubble-diagram adjacency list from interior-door connectivity. We filter out samples with disconnected graphs or invalid polygons. Full preprocessing details appear in Appendix \ref{sec:rplan_conversion}.

\section{Our Method}

To enable the model to generate valid floor plans from explicit design specifications, we employ a two-stage training procedure: supervised fine-tuning in stage one, and reinforcement learning–based fine-tuning in the second stage using GRPO with rewards derived from automated quantitative metrics, together with a hard feasibility condition that assigns zero reward to invalid or overlapping outputs. 

For all experiments, we use Llama-3.3-70B-Instruct as the backbone \citep{grattafiori2024llama3}. At inference time, we use \textbf{best-of-10 sampling}, selecting for each prompt the candidate with the smallest overlap area and breaking ties by Compatibility. Complete technical training and inference details appear in Appendix~\ref{sec:training_details}.

\subsection{Supervised Fine-Tuning}
In the first phase, we perform supervised instruction tuning so that the model learns to translate structured prompts into JSON-encoded floor plans. Let $x=\{(k_i,v_i)\}_{i=1}^K$ be the conditioning input (e.g., room count, total area, bubble diagram), and let $y_{1:T}$ be the ground-truth token sequence. We adapt the pre-trained LLM by minimizing the negative log-likelihood:

\begin{align}
    L^{\mathrm{SFT}}(\theta)
    =
    \mathbb{E}_{(x,y)\sim\mathcal{D}}
    \left[
    -\sum_{t=1}^{T} \log \pi_{\theta}(y_t \mid y_{<t}, x)
    \right].
    \label{eq:sft}
\end{align}

This objective drives the model to predict each token in \(y\) accurately given its prefix and the full specification \(x\), producing floor plans that satisfy numerical and connectivity constraints. 

After supervised fine-tuning, the LLM generates plausible floor plans but still produces overlapping polygons, a limitation that, to our knowledge, is not explicitly addressed by the methods we compare against and that must be addressed before deployment. We therefore introduce a second training stage to mitigate this issue.

\begin{table*}[t]
  \centering
  \setlength{\tabcolsep}{3.5pt} 
  \begin{tabular}{c|cccc|c|cccc}
    \toprule
    Model
      & \multicolumn{4}{c|}{Compatibility↓}
      & Realism↑
      & \multicolumn{4}{c}{Diversity↓} \\
    \cmidrule(lr){1-1} \cmidrule(lr){2-5} \cmidrule(lr){6-6} \cmidrule(lr){7-10}
    Task
      & 5 & 6 & 7 & 8
      & 8
      & 5 & 6 & 7 & 8 \\
    \midrule
    
    ~\cite{ashual2019specifying}
      & 7.5   & 9.2   & 10.0  & 11.8
      & -1.00
      & 120.6 & 172.5 & 162.1 & 183.0 \\
    ~\cite{johnson2018image}
      & 7.7   & 6.5   & 10.2  & 11.3
      & -1.00
      & 167.2 & 168.4 & 186.0 & 186.0 \\
    
    House‑GAN~\cite{nauata2020house}
      & 2.5 & 2.4 & 3.2 & 5.3
      & -0.95
      & 37.5 & 41.0 & 32.9 & 66.4 \\
    
    House‑GAN++~\cite{nauata2021house}
      & 1.9 & 2.2 & 2.4 & 3.9
      & -0.52
      & 30.4 & 37.6 & 27.3 & 32.9 \\

    HouseDiffusion~\cite{shabani2023housediffusion}
      & 1.5 & 1.2 & 1.7 & 2.5
      & -0.19
      & 11.2 & 10.3 & 10.4 & 9.5 \\
      
    \midrule
    \textbf{(Ours)}
      & \textbf{0.01} & \textbf{0.02} & \textbf{0.10} & \textbf{0.15}
      & \textbf{0.03}
      & \textbf{9.0 }& \textbf{8.8} & \textbf{7.8} & \textbf{7.0} \\
    \bottomrule
  \end{tabular}
  \caption{Main quantitative results comparing our approach with previous methods on Compatibility (↓), Realism (↑), and Diversity (↓). Our method fine-tunes a Llama-3.3-70B-Instruct in two stages: first supervised fine-tuning, followed by reinforcement learning with verifiable rewards. For our method, results are reported using best-of-10 sampling, selecting the candidate with the smallest overlap area and, in case of ties, the lowest Compatibility.}
  \label{tab:gen_performance}
\end{table*}

\subsection{GRPO}

In the second phase of training, we perform reinforcement-learning-based fine-tuning using reinforcement learning with verifiable rewards (RLVR). Specifically, we use Group Relative Policy Optimization (GRPO) as our RLVR algorithm \citep{grpo}. GRPO is a PPO-style policy optimization method \citep{schulman2017ppo} that improves the policy using relative reward comparisons within groups of sampled outputs, rather than relying on a learned value function.

For each input specification $x$, we sample a group of $G=4$ candidate floor plans $\{y_i\}_{i=1}^{G}$ from the current policy. Each candidate is assigned a scalar reward $R(x, y_i)$ based on automatically computed constraint adherence metrics. We then normalize rewards within the sampled group to obtain group-relative advantages:
\begin{equation}
\hat{A}_i
=
\frac{R(x,y_i)-\mu_x}{\sigma_x+\epsilon},
\label{eq:grpo-advantage}
\end{equation}
where $\mu_x$ and $\sigma_x$ denote the mean and standard deviation of rewards over the sampled group for input $x$, and $\epsilon$ is a small constant for numerical stability.

The policy is then updated using a PPO-style surrogate objective:
\begin{equation}
L^{\mathrm{GRPO}}(\theta)
=
\mathbb{E}_{x\sim\mathcal{D}}
\left[
\frac{1}{G}
\sum_{i=1}^{G}
\frac{\pi_\theta(y_i \mid x)}{\pi_{\theta_{\mathrm{old}}}(y_i \mid x)}
\hat{A}_i
\right].
\label{eq:grpo-obj}
\end{equation}

Intuitively, this objective increases the probability of candidates that perform better than the group average and decreases the probability of worse-performing candidates. By comparing outputs relative to one another for the same prompt, GRPO provides a stable training signal without requiring a separate critic network.

In our setting, the model generates multiple candidate floor plans for each input prompt. Any candidate that fails to parse as valid JSON or contains overlapping room polygons receives a reward of $0$. Overlap is therefore handled as a hard feasibility condition rather than as a separate reward term. For each candidate $y$ that parses as valid JSON and is free of overlapping room polygons, we compute two automatically verifiable reward terms:

\begin{enumerate}
    \item \textbf{Connectivity Reward:} We reconstruct a connectivity graph from the generated layout and compare it with the input connectivity graph. The reward increases when the generated floor plan better matches the requested connectivity structure, and decreases when connectivity errors are present. This yields a reward $r_{\mathrm{conn}}(y) \in [0,1]$, where higher is better, and $1$ indicates an exact match.

    \item \textbf{Total Area Reward:} Let $A(y)$ denote the total area of the generated layout for candidate $y$, and let $A^\star$ denote the target total area specified in the input prompt. We define the total area error as:
    \begin{equation}
        \mathrm{TAE}(y) = \frac{|A(y) - A^\star|}{A^\star},
    \end{equation}
    and the corresponding reward as:
    \begin{equation}
        r_{\mathrm{TA}}(y) = \max(0, 1 - \mathrm{TAE}(y)).
    \end{equation}

\end{enumerate}

The final scalar reward is the equally weighted average of the Connectivity Reward and Total Area Reward. We found equal weighting to work well empirically, and leave a more systematic study of alternative reward weightings to future work.

\section{Evaluation}

We follow the evaluation protocol of previous work \citep{nauata2020house,nauata2021house,shabani2023housediffusion} and divide all floor plan samples into four groups based on the number of rooms (5, 6, 7, or 8 rooms). For each experiment, one group is held out completely, while the model is trained using 100\% of the remaining three groups. The held-out group is split into two equal parts: 50\% for validation and 50\% for testing. This specific validation–test split is not explicitly defined in the original protocol. Still, we adopt it to provide a separate validation set for monitoring performance during training and a dedicated test set for final reporting. 

For example, when performing the five-room task, all five-room samples are removed from the training set, and the model is trained on the six-, seven-, and eight-room samples. The five-room group is then split evenly for validation and testing. Following the evaluation protocol, all experiments are tested using a random subset of 1,000 samples from the test portion of the held-out group. This setup ensures that the model must generalize to unseen configurations rather than memorizing layouts for a specific room count.

We compare our method against bubble diagram constrained floor plan generators, including \textit{House-GAN} \cite{nauata2020house}, \textit{House-GAN++} \cite{nauata2021house}, and \textit{HouseDiffusion} \cite{shabani2023housediffusion}, where \textit{HouseDiffusion} represents the current peer-reviewed state of the art. We also compare against scene graph constrained image generation methods \cite{johnson2018image,ashual2019specifying}.

\begin{table*}[t]
  \centering
  \setlength{\tabcolsep}{3.25pt}
  \begin{tabular}{c|c|cccccc}
    \toprule
    Experiment & Task & Room Area↓ & Room ID↑ & Overlap↓ & \% Overlap↓ & Compatibility↓ & Diversity↓    \\
    \midrule
    \textcolor[HTML]{C4A484}{\textbf{Few‑shot}}             & \multirow{3}{*}{5}&0.27$\pm$0.22&0.96$\pm$0.19&0.55$\pm$0.50&0.07$\pm$0.10&2.93$\pm$1.26&45.96$\pm$0.00\\
    \textcolor[HTML]{FF6AD5}{\textbf{SFT}} && \textbf{0.10$\pm$0.08} & 1.00$\pm$0.00 & 0.12$\pm$0.33 & 0.00$\pm$0.02 & 0.02$\pm$0.16 & \textbf{8.60$\pm$0.00}\\
    \textcolor[HTML]{FF6AD5}{\textbf{SFT}} + \textcolor[HTML]{966BFF}{\textbf{RLVR}} && 0.11$\pm$0.08 & \textbf{1.00$\pm$0.00} & \textbf{0.03$\pm$0.17} & \textbf{0.00$\pm$0.00} & \textbf{0.01$\pm$0.14} & 8.96$\pm$0.00 \\
    \cmidrule(lr){1-8}
    \textcolor[HTML]{C4A484}{\textbf{Few‑shot}}               & \multirow{3}{*}{6}&0.19$\pm$0.19&1.00$\pm$0.00&0.54$\pm$0.50&0.06$\pm$0.09&4.27$\pm$1.47&40.09$\pm$0.00\\
    \textcolor[HTML]{FF6AD5}{\textbf{SFT}} && \textbf{0.10$\pm$0.07} & 1.00$\pm$0.00 & 0.14$\pm$0.35 & 0.00$\pm$0.02 & 0.04$\pm$0.23 & \textbf{7.59$\pm$0.00} \\
    \textcolor[HTML]{FF6AD5}{\textbf{SFT}} + \textcolor[HTML]{966BFF}{\textbf{RLVR}} && 0.12$\pm$0.08 & \textbf{1.00$\pm$0.00} & \textbf{0.05$\pm$0.21} & \textbf{0.00$\pm$0.01}& \textbf{0.02$\pm$0.17} & 8.79$\pm$0.00 \\
    \cmidrule(lr){1-8}
    \textcolor[HTML]{C4A484}{\textbf{Few‑shot}}               & \multirow{3}{*}{7}&0.15$\pm$0.14&0.99$\pm$0.05&0.50$\pm$0.50&0.05$\pm$0.08&5.27$\pm$1.19&41.73$\pm$0.00\\
    \textcolor[HTML]{FF6AD5}{\textbf{SFT}} && \textbf{0.09$\pm$0.06} & 1.00$\pm$0.00 & 0.23$\pm$0.42 & 0.01$\pm$0.02 & 0.17$\pm$0.51 & \textbf{6.79$\pm$0.00} \\
    \textcolor[HTML]{FF6AD5}{\textbf{SFT}} + \textcolor[HTML]{966BFF}{\textbf{RLVR}} && 0.12$\pm$0.07 & \textbf{1.00$\pm$0.00} & \textbf{0.09$\pm$0.29} & \textbf{0.00$\pm$0.01}& \textbf{0.10$\pm$0.40} & 7.79$\pm$0.00 \\
    \cmidrule(lr){1-8}
    \textcolor[HTML]{C4A484}{\textbf{Few‑shot}}               & \multirow{3}{*}{8} &0.12$\pm$0.10&0.91$\pm$0.20&0.51$\pm$0.50&0.04$\pm$0.06&6.89$\pm$0.71&49.84$\pm$0.00\\
    \textcolor[HTML]{FF6AD5}{\textbf{SFT}} && \textbf{0.08$\pm$0.05} & 1.00$\pm$0.00 & 0.37$\pm$0.48 & 0.01$\pm$0.03 & 0.41$\pm$0.73 & \textbf{6.44$\pm$0.00} \\
    \textcolor[HTML]{FF6AD5}{\textbf{SFT}} + \textcolor[HTML]{966BFF}{\textbf{RLVR}} && 0.10$\pm$0.06 & \textbf{1.00$\pm$0.00} & \textbf{0.13$\pm$0.33} & \textbf{0.00$\pm$0.01}& \textbf{0.15$\pm$0.48} & 6.96$\pm$0.00 \\
    \bottomrule
  \end{tabular}
  \caption{Task-wise results for Llama-3.3-70B-Instruct under \textcolor[HTML]{C4A484}{\textbf{few-shot}} with three examples, \textcolor[HTML]{FF6AD5}{\textbf{supervised fine-tuning}}, and \textcolor[HTML]{FF6AD5}{\textbf{supervised fine-tuning}} plus \textcolor[HTML]{966BFF}{\textbf{reinforcement learning with verifiable rewards}}. \textcolor[HTML]{FF6AD5}{\textbf{SFT}} + \textcolor[HTML]{966BFF}{\textbf{RLVR}} yields the lowest Overlap, \% Overlap, and Compatibility across all tasks, while maintaining similar Room Area performance. All results in this table use best-of-10 sampling, with selection based on the smallest overlap area and, in case of ties, the lowest Compatibility.}
  \label{tab:task_breakdown}
\end{table*}

\subsection{Existing Metrics}

We evaluate our method using three existing metrics: \textbf{Compatibility}, \textbf{Realism}, and \textbf{Diversity}, following prior work \citep{nauata2020house,nauata2021house,shabani2023housediffusion}. \textbf{Compatibility} is computed as the graph edit distance \citep{sanfeliu1983distance} between the input bubble diagram and the bubble diagram reconstructed from the generated floor plan JSON. \textbf{Realism} is measured through feedback from a set of volunteers: each person views 10 randomized pairs of layouts (one ground truth and one generated) and indicates which layout appears more realistic, or whether both appear equally realistic. \textbf{Diversity} is measured using the Fréchet Inception Distance (FID) \citep{heusel2017gans}, which compares the feature distributions of generated floor plan images and ground truth floor plan images. A more detailed description of these three metrics is provided in Appendix \ref{sec:metrics_appendix}.

\subsection{Our Metrics}

We also introduce four metrics that directly evaluate adherence to prompt-specified constraints and quantify polygon overlaps. To our knowledge, this is the first work to report metrics that explicitly measure both overlaps and room-size constraints. For all four metrics, we report the mean and standard deviation across samples.

\begin{itemize}
    \item \textbf{Room Area $\downarrow$:} Mean absolute percentage error of per-room area with respect to the prompt. 
    \item \textbf{Room ID $\uparrow$:} Exact-match accuracy of \texttt{id} relative to the prompt. Each \texttt{id} encodes both room type and instance index (e.g., "bedroom|1").
    \item \textbf{Overlap $\downarrow$:} Boolean indicator of whether any generated room polygons overlap. 
    \item \textbf{\% Overlap $\downarrow$:} Total overlapped area divided by the generated total area.
\end{itemize}

\section{Experiments}

\subsection{Quantitative Evaluations}
Table \ref{tab:gen_performance} reports the main quantitative results; for completeness, Appendix~\ref{sec:full_gen_performance} reproduces this table with standard deviations.  Baseline numbers are taken from \textit{HouseDiffusion} \cite{shabani2023housediffusion}. Our method achieves the lowest Compatibility score across all tasks (0.01 to 0.15), indicating near-perfect alignment with the input bubble diagram. For Diversity, it also attains the lowest values (7.0 to 9.0). On the eight-room task, relative to \textit{HouseDiffusion}, our method reduces Compatibility by \textbf{94.00\%} (from 2.50 to 0.15) and improves Diversity by \textbf{26.32\%} (from 9.5 to 7.0). 

Table \ref{tab:task_breakdown} reports our metrics across training settings.  After the second training stage (SFT + RLVR), which is our best-performing model, Room Area remains between 0.10 and 0.12, meaning the average per-room area deviates by only about 10\% to 12\% from the requested values and does not degrade as task complexity increases. Room ID is 1.00, indicating perfect labeling of both room type and instance-level identifiers (as specified in the prompt) across all tasks. However, Overlap increases as the room count increases (0.03, 0.05, 0.09, 0.13). The \% Overlap is near zero across all tasks, indicating that, for a typical generated floor plan, the overlapped area is only a very small fraction of that plan’s total area.

\begin{table*}[t]
  \centering
  \setlength{\tabcolsep}{2.5pt}
  \begin{tabular}{c|c|c c c c c c}
    \toprule
    Experiment                      & Task & Budget & Room Area↓  & Room ID↑ & Overlap↓    & \% Overlap↓            & Compatibility↓ \\
    \midrule
    \multirow{3}{*}{\shortstack[c]{%
      \textcolor[HTML]{FF6AD5}{\textbf{SFT}} 
      $+$
      \textcolor[HTML]{966BFF}{\textbf{RLVR}}%
    }}
                               & \multirow{3}{*}{8}
                                     & 1                & $\mathbf{0.09\pm0.05}$ & $1.00\pm0.00$ & $0.26\pm0.44$ & $0.01\pm0.02$ & $1.89\pm1.97$                        \\
                               &  & 10  & $0.10\pm0.06$ & $1.00\pm0.00$ & $0.13\pm0.33$      & $0.00\pm0.01$ & $0.15\pm0.48$                          \\
                               &  & 100 & $0.09\pm0.06$ & $\mathbf{1.00\pm0.00}$ & $\mathbf{0.10\pm0.30}$ & $\mathbf{0.00\pm0.01}$ & $\mathbf{0.02\pm0.16}$ \\
    \bottomrule
  \end{tabular}
  \caption{Effect of generation budget \(n\) on the eight-room task for Llama-3.3-70B-Instruct after two-stage fine-tuning. For each prompt, we sample \(n \in \{1,10,100\}\) candidates and select the one with the least overlap, breaking ties by Compatibility.}
  \label{tab:budget_effects}
\end{table*}

\begin{figure*}[p]
    \centering
    \includegraphics[width=0.90\textwidth]{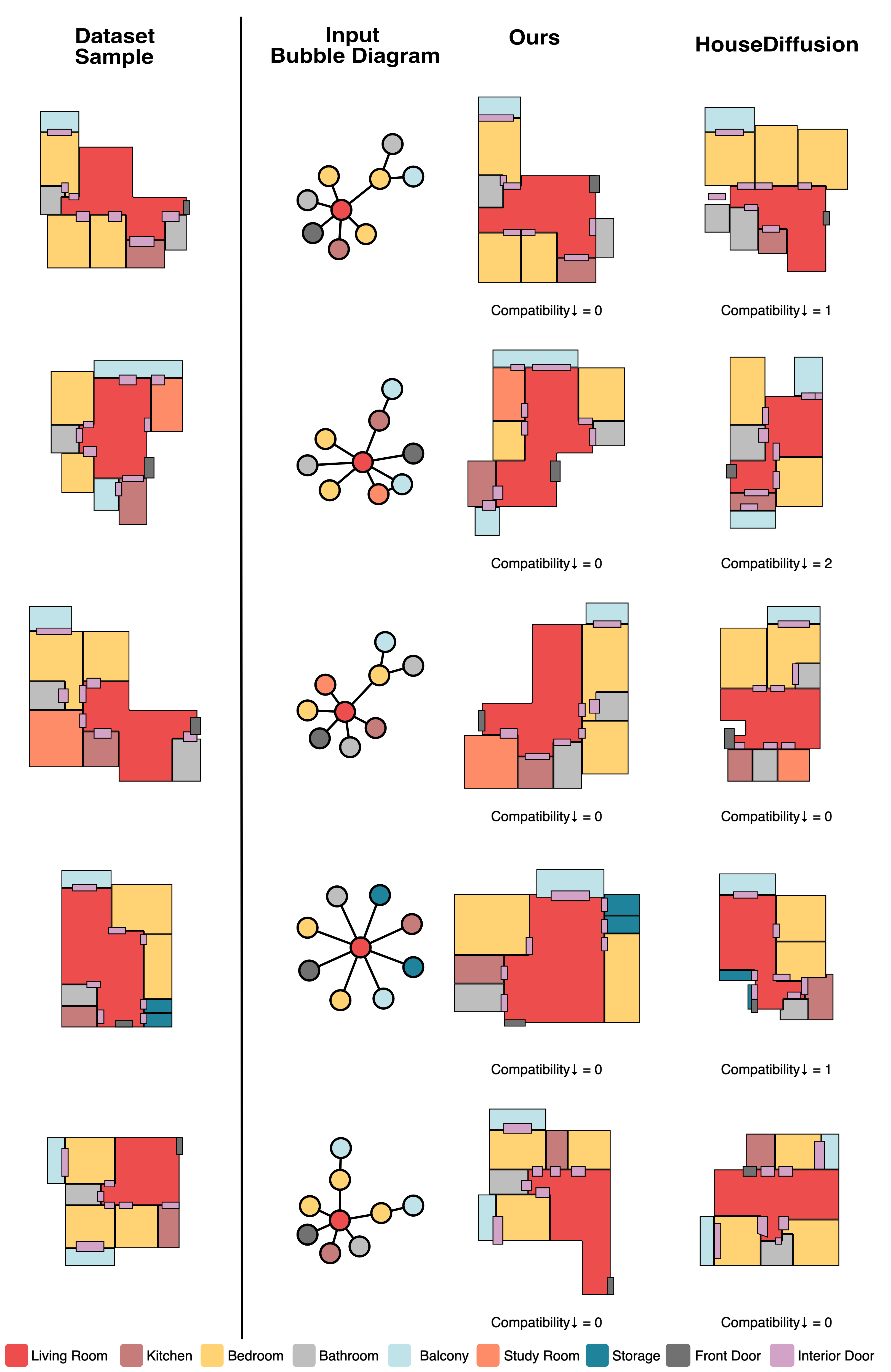}
    \caption{Qualitative comparison against \textit{HouseDiffusion}. We replicate \textit{HouseDiffusion}'s visualization to enable direct visual comparison. From left to right: dataset sample (reference), input bubble diagram, and layouts generated by our method and \textit{HouseDiffusion} from the same bubble diagram. Numbers under the outputs report Compatibility (graph edit distance; lower is better). Our method better matches the reference room shapes and per-room proportions while achieving equal or lower Compatibility across the shown examples.}
    \label{fig:visual_comparison}
\end{figure*}

\subsection{Qualitative Evaluations}
Figure \ref{fig:visual_comparison} compares our method with \textit{HouseDiffusion} for the same input bubble diagram and shows a reference dataset sample. Across the five examples, our method matches the requested connectivity in every case, preserving the prompt-specified bubble diagram more reliably (i.e., showing a lower Compatibility score).

Beyond connectivity, our method produces floor plans whose room shapes and per-room proportions more closely match the reference samples. 
For instance, in the third row of Figure \ref{fig:visual_comparison}, the study, kitchen, and bathroom connected to the living room all have sizes similar to those in the reference.

We also verified this empirically by collecting feedback from a set of volunteers following the evaluation setup outlined in \textit{HouseDiffusion}.
The feedback setup included 10 pairwise comparisons and one warm-up example. Each comparison showed a ground-truth floor plan and a floor plan generated by our method for the same bubble diagram.
The order of comparisons and the position of the ground-truth and generated floor plans were randomized (e.g., sometimes the GT is on the left side, sometimes on the right).
We asked a set of 40 volunteers who did not know about the project and had no prior experience designing floor plans to provide feedback.
Volunteers were asked which layout appeared more realistic: (a)  the ground truth (-1 point), (b) the generated one (+1 point), (c) both equally (0 points).
The Realism score in Table \ref{tab:gen_performance} was calculated as the mean point score across all comparisons and volunteers.
A screenshot of the feedback interface and the instructions given to volunteers is provided in Appendix \ref{sec:realism-survey}.

Based on feedback from $40$ volunteers, our model achieves a \textbf{Realism score of 0.028}, a value very close to zero that indicates the generated layouts were judged similarly to the ground truth on average. This compares favorably with the Realism scores reported in prior work (see Table \ref{tab:gen_performance}; this value is shown as 0.03 in the table after rounding to two decimal places).

\subsection{Effect of the Training Stages}

Table \ref{tab:task_breakdown} compares three settings for Llama-3.3-70B-Instruct: few-shot prompting, supervised fine-tuning, and supervised fine-tuning followed by reinforcement learning with verifiable rewards. Few-shot prompting with three examples serves only as a baseline. It is unstable and scales poorly with task complexity: Compatibility rises from 2.93 in the five-room task to 6.89 in the eight-room task, while Overlap remains around 0.50 to 0.55 with high variance, even under best-of-10 selection. The high Diversity scores reflect uncontrolled variability, as many samples violate the bubble diagram, contain overlaps, or both.

\begin{figure}
    \centering
    \includegraphics[width=0.45\textwidth]{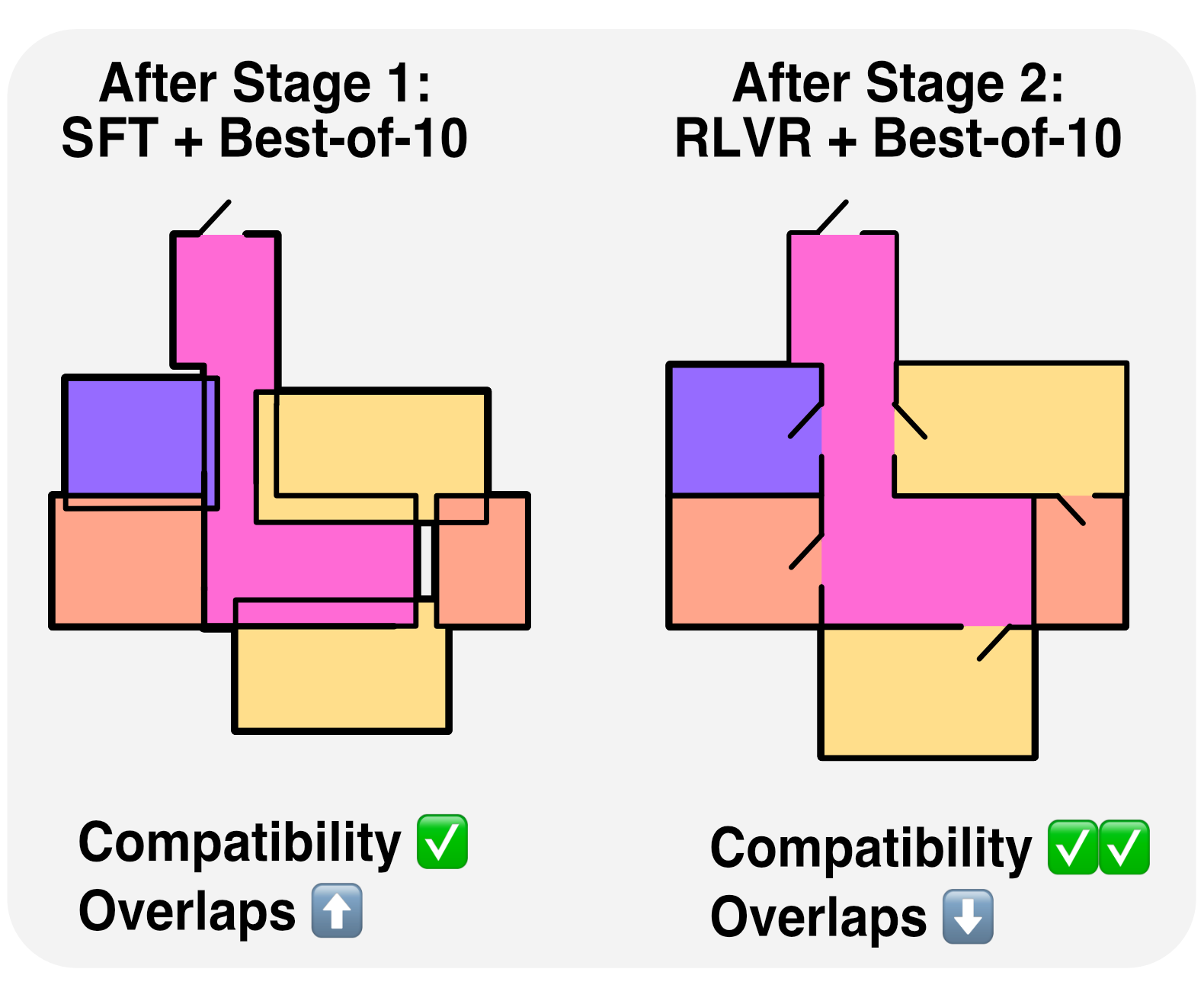}
    \caption{\textbf{Illustration of Post-Training Results.} Both Stage 1 (supervised fine-tuning, SFT) and Stage 2 (reinforcement learning with verifiable rewards, RLVR) use best-of-10 sampling. After Stage 1, the model generally follows the input bubble diagram but can still produce overlapping polygons. Stage 2 improves Compatibility and reduces Overlap.}
    \label{fig:effect-training-stages}
\end{figure}

After supervised fine-tuning, the model learns the mapping from structured prompts to JSON floor plan and already achieves low Room Area with perfect Room ID. However, as the room count increases, overlaps and connectivity mismatches also increase. Averaged across the five- to eight-room tasks, adding RLVR reduces Overlap by 65\% and Compatibility by 56\% relative to supervised fine-tuning. Meanwhile, \% Overlap remains near zero, Room Area stays between 0.10 and 0.12, and Room ID remains perfect, indicating that these gains do not come at the expense of size accuracy or room labeling. Figure~\ref{fig:effect-training-stages} highlights this progression across training stages.

\begin{table*}[t]
  \centering
  \setlength{\tabcolsep}{2.5pt}
  \renewcommand{\arraystretch}{1.1}
  \begin{tabular}{m{1.8cm}| c c c c c c}
    \toprule
    \centering Input Type & Room Area$\downarrow$ & Room ID$\uparrow$ & Overlap$\downarrow$ & \% Overlap$\downarrow$ & Compatibility$\downarrow$ & Diversity$\downarrow$ \\
    \midrule
    \centering JSON
    & $0.10\pm0.06$
    & $\mathbf{1.00\pm0.00}$
    & $\mathbf{0.13\pm0.33}$
    & $\mathbf{0.00\pm0.01}$
    & $\mathbf{0.15\pm0.48}$
    & $\mathbf{6.96\pm0.00}$ \\
    \centering Natural-Language
    & $0.10\pm0.06$
    & $\mathbf{1.00\pm0.00}$
    & $0.14\pm0.35$
    & $\mathbf{0.00\pm0.01}$
    & $0.37\pm0.85$
    & $7.37\pm0.00$ \\
    \bottomrule
  \end{tabular}
  \caption{Comparison between structured JSON and natural-language inputs on the eight-room task for our best-performing model (\textcolor[HTML]{FF6AD5}{\textbf{SFT}} $+$ \textcolor[HTML]{966BFF}{\textbf{RLVR}}) using best-of-10 sampling. We fine-tune the model only on JSON inputs, while the natural-language inputs come from deterministic verbalizations of the same constraints using three templates.}
  \label{tab:nl_generalization}
\end{table*}

\subsection{Effect of the Generation Budget}
\label{sec:generation_budget}

We study the effect of the generation budget $n$ on the eight-room task using best-of-$n$ selection. For each prompt, we sample \(n\) candidates with temperature 0.7 and top-p 0.9, then choose the candidate with the smallest overlap area, using Compatibility to break ties. Table \ref{tab:budget_effects} reports results for \(n \in \{1, 10, 100\}\). Increasing the budget from 10 to 100 reduces Overlap by 23.1\% and Compatibility by 86.7\%, while \% Overlap remains at \(0.00 \pm 0.01\). However, sampling 100 candidates per prompt substantially increases inference cost.

\subsection{Zero-Shot Generalization to Natural-Language Constraints}

Although our primary interface is structured JSON, we also evaluate whether the same model can handle natural-language interaction. To do so, we compare two input types: the original JSON constraints and natural-language descriptions obtained through deterministic verbalizations of the same information. To introduce limited phrasing variation without changing the meaning, we use three verbalization templates. The model is fine-tuned only on JSON inputs and is not retrained on natural-language prompts.

Table~\ref{tab:nl_generalization} shows results on the eight-room task for our best-performing model (SFT + RLVR) using best-of-10 sampling. Using natural-language inputs instead of JSON leads to almost no change in Room Area, Room ID, Overlap, or \% Overlap. The main difference is in Compatibility, which rises from $0.15 \pm 0.48$ with JSON inputs to $0.37 \pm 0.85$ with natural-language inputs. One possible explanation is that connectivity constraints are more sensitive to wording, so verbalized descriptions introduce some ambiguity that is not present in the structured format. At the same time, the model’s overall stability across the two input types suggests that it still retains the ability to interpret free-form constraint descriptions.

Overall, these results show that the model can handle natural-language inputs in a zero-shot setting for this task, while structured JSON remains the more reliable interface when precise connectivity control is required.

\section{Conclusions}

We introduced a structured generator that maps a bubble diagram and numerical specifications to a floor plan. Our two-stage training recipe, which combines supervised fine-tuning with reinforcement learning with verifiable rewards, trains an LLM to improve Compatibility while reducing overlap between room polygons and preserving area-related accuracy. Across the five- to eight-room tasks, our model achieves the lowest Compatibility (0.01 to 0.15), the lowest Diversity values (7.0 to 9.0), and a Realism score of 0.028, indicating that the generated layouts were judged similarly to the ground truth on average according to volunteer feedback. In the most complex eight-room setting, our method reduces Compatibility by 94\% and improves Diversity by 26\% relative to the peer-reviewed state of the art, \textit{HouseDiffusion}. Together, these results show that text-based models can produce controllable, CAD-ready floor plans under explicit connectivity constraints and structured area specifications.

Limitations include occasional overlaps at higher room counts, inference requires multiple samples per prompt (increasing computational cost), and the use of single-floor residential floor plans from Asia derived from \textit{RPLAN}, potentially limiting generalizability to other architectural styles or building types. Future work will scale to multi-floor residential buildings, add additional verifiable objectives to the reward (for example, circulation and daylight proxies), and enable interactive editing while preserving constraints. More broadly, the same structured approach should transfer to other graph and schema-constrained design problems beyond floor plan design.

\section*{Limitations}

Our reinforcement learning with verifiable rewards (RLVR) optimizes only a limited set of automatically checkable objectives. These rewards are necessary but not sufficient for architectural validity and usability: many constraints that determine real-world quality (e.g., circulation, egress, accessibility, daylight, structural constraints, and local building codes) are not modeled or verified. Consequently, strong performance on our metrics should not be interpreted as evidence of code compliance or construction readiness.

Our experiments are conducted on floor plans derived from RPLAN, which contains single-floor residential layouts from a specific geographic and cultural distribution. This scope limits generalization to other regions, architectural styles, multi-family housing, commercial programs, or multi-floor buildings. In addition, the conversion pipeline from rasterized annotations to metric polygons may introduce noise (e.g., imperfect polygon reconstruction or scaling). Our results implicitly assume these artifacts are not dominant; performance may degrade on datasets with different annotation conventions, room taxonomies, higher geometric complexity, or more complex constraint sets.

Finally, our empirical evidence is restricted to a single underlying dataset and a small set of task regimes (held-out room-count generalization). Although we evaluate on 1{,}000 test samples per task, we do not provide extensive ablations across multiple random seeds, alternative backbones, or alternative reward weightings, and we do not provide an extensive study of sensitivity to inference-time compute budgets beyond Section~\ref{sec:generation_budget}.

\section*{Acknowledgments}

We thank the P2 program at Mila – Quebec AI Institute, Mila’s IDT team for technical support, CIFAR for support through the Canada CIFAR AI Chair program, and NSERC for support under the Discovery Grants program.

\bibliography{custom}

\appendix

\section{Appendix}

\subsection{JSON Data Schemas for Floor Plan Generation}
\label{sec:json_schemas}

In this section, we describe the exact JSON schemas our model consumes and produces. Each schema is built around a top-level \texttt{spaces} array whose elements represent any floor plan entity, including both rooms and doors (interior or front). 

In the input schema, each space must include an \texttt{id} and a \texttt{room\_type} label, and may specify either an explicit \texttt{area} for irregular-polygon shapes or a pair of \texttt{height} and \texttt{width} values for rectangular spaces. The input also includes a \texttt{room\_count}, a \texttt{total\_area} constraint, and an \texttt{input\_graph} dictionary encoding the bubble diagram; front doors must be specified explicitly in the \texttt{spaces} array, while interior door connections are inferred from the bubble diagram. The input schema is summarized in Table~\ref{tab:input_data_structure}. 

In the output schema, each space object includes its computed \texttt{area} and a \texttt{floor\_polygon} array of vertices defining its precise footprint in absolute coordinates. All area values (\texttt{area}, \texttt{total\_area}) are given in square meters. The output schema is summarized in Table~\ref{tab:output_data_structure}.

\begin{table}[t]
  \centering
  \small
  \setlength{\tabcolsep}{1mm}
  \begin{tabular}{@{}p{0.35\columnwidth}@{}p{0.65\columnwidth}@{}}
    \toprule
    \textbf{Field}       & \textbf{Description} \\
    \midrule
    \texttt{room\_count} & Total number of rooms \\
    \texttt{total\_area} & Sum of all room areas \\
    \texttt{spaces}      & Array of space objects \\
    \hspace{4mm}\texttt{id}         & Unique identifier (e.g., \texttt{"bedroom\textbar0"}) \\
    \hspace{4mm}\texttt{room\_type} & Semantic label (e.g., \texttt{"bedroom"}) \\
    \hspace{4mm}\texttt{area}       & Area for an irregular polygon space (omit \texttt{height} and \texttt{width}) \\
    \hspace{4mm}\texttt{height}     & Height of bounding rectangle for a regular polygon space (omit \texttt{area}) \\
    \hspace{4mm}\texttt{width}      & Width of bounding rectangle for a regular polygon space (omit \texttt{area}) \\
    \texttt{input\_graph} & Bubble diagram. Each key is a space ID mapping to an array of its neighbor IDs \\
    \bottomrule
  \end{tabular}
  \caption{Input JSON data structure for floor plan generation.}
  \label{tab:input_data_structure}
\end{table}

\begin{table}[t]
  \centering
  \small
  \setlength{\tabcolsep}{1mm}
  \begin{tabular}{@{}p{0.35\columnwidth}@{}p{0.65\columnwidth}@{}}
    \toprule
    \textbf{Field}           & \textbf{Description} \\
    \midrule
    \texttt{room\_count}     & Total number of rooms in the generated floor plan \\
    \texttt{total\_area}     & Sum of all generated room areas \\
    \texttt{spaces}      & Array of space objects \\
    \hspace{4mm}\texttt{id}           & Unique identifier (e.g., \texttt{"bedroom\textbar0"}) \\
    \hspace{4mm}\texttt{room\_type}   & Semantic label (e.g., \texttt{"bedroom"}) \\
    \hspace{4mm}\texttt{area}         & Area of the polygon space \\
    \hspace{4mm}\texttt{floor\_polygon} & List of vertices outlining the space polygon \\
    \hspace{8mm}\texttt{x}            & X–coordinate \\
    \hspace{8mm}\texttt{y}            & Y–coordinate \\
    \bottomrule
  \end{tabular}
  \caption{Output JSON data structure for generated floor plans.}
  \label{tab:output_data_structure}
\end{table}

\begin{table*}[t]
  \centering
  \setlength{\tabcolsep}{4pt} 
  \begin{tabular}{c|cccc}
    \toprule
    Model & \multicolumn{4}{c}{Compatibility↓} \\
    \cmidrule(lr){1-1} \cmidrule(lr){2-5}
    Task  & 5 & 6 & 7 & 8 \\
    \midrule

    ~\cite{ashual2019specifying}
      & 7.5$\pm$0.0   & 9.2$\pm$0.0   & 10.0$\pm$0.0  & 11.8$\pm$0.0 \\
    ~\cite{johnson2018image}
      & 7.7$\pm$0.0   & 6.5$\pm$0.0   & 10.2$\pm$0.0  & 11.3$\pm$0.1 \\
    
    House‑GAN~\cite{nauata2020house}
      & 2.5$\pm$0.1 & 2.4$\pm$0.1 & 3.2$\pm$0.0 & 5.3$\pm$0.0 \\
    
    House‑GAN++~\cite{nauata2021house}
      & 1.9$\pm$0.3 & 2.2$\pm$0.3 & 2.4$\pm$0.3 & 3.9$\pm$0.5 \\

    HouseDiffusion~\cite{shabani2023housediffusion}
      & 1.5$\pm$0.0 & 1.2$\pm$0.0 & 1.7$\pm$0.0 & 2.5$\pm$0.0 \\
      
    \midrule
    \textbf{(Ours)}
      & \textbf{0.01$\pm$0.1} & \textbf{0.02$\pm$0.2} & \textbf{0.10$\pm$0.4} & \textbf{0.15$\pm$0.5} \\
    \bottomrule
  \end{tabular}
  \caption{Compatibility (↓) results (mean $\pm$ std) across tasks.}
  \label{tab:compatibility_only}
\end{table*}

\begin{table*}[t]
  \centering
  \setlength{\tabcolsep}{4pt} 
  \begin{tabular}{c|c|cccc}
    \toprule
    Model & Realism↑ & \multicolumn{4}{c}{Diversity↓} \\
    \cmidrule(lr){1-1} \cmidrule(lr){2-2} \cmidrule(lr){3-6}
    Task  & 8 & 5 & 6 & 7 & 8 \\
    \midrule
    
    ~\cite{ashual2019specifying}
      & -1.00
      & 120.6$\pm$0.5 & 172.5$\pm$0.2 & 162.1$\pm$0.4 & 183.0$\pm$0.4 \\
    ~\cite{johnson2018image}
      & -1.00
      & 167.2$\pm$0.3 & 168.4$\pm$0.4 & 186.0$\pm$0.4 & 186.0$\pm$0.4 \\
    
    House‑GAN~\cite{nauata2020house}
      & -0.95
      & 37.5$\pm$1.1 & 41.0$\pm$0.6 & 32.9$\pm$1.2 & 66.4$\pm$1.7 \\
    
    House‑GAN++~\cite{nauata2021house}
      & -0.52
      & 30.4$\pm$4.4 & 37.6$\pm$3.3 & 27.3$\pm$4.9 & 32.9$\pm$4.9 \\

    HouseDiffusion~\cite{shabani2023housediffusion}
      & -0.19
      & 11.2$\pm$0.2 & 10.3$\pm$0.2 & 10.4$\pm$0.4 & 9.5$\pm$0.1 \\
      
    \midrule
    \textbf{(Ours)}
      & \textbf{0.03}
      & \textbf{9.0$\pm$0.0} & \textbf{8.8$\pm$0.0} & \textbf{7.8$\pm$0.0} & \textbf{7.0$\pm$0.0} \\
    \bottomrule
  \end{tabular}
  \caption{Realism (↑) and Diversity (↓) results (mean $\pm$ std) across tasks.}
  \label{tab:realism_diversity_only}
\end{table*}

\subsection{RPLAN Conversion}
\label{sec:rplan_conversion}

\textbf{RPLAN} \citep{rplan} is a manually collected dataset of 80,788 real-world floor plans of buildings in Asia. Each floor plan in \textit{RPLAN} is stored as a $256\times256\times4$ four-channel image. Channels 1 and 2 store interior and exterior boundary information; channel 3 contains room information where each pixel value denotes which room it belongs to; channel 4 has extra information to distinguish rooms with the same room type value in channel 3. 

To convert this 4-channel image into a JSON structure, we first use the same data-reader as in \textit{House-GAN++}\footnote{\url{https://github.com/sepidsh/Housegan-data-reader}}, then process each entry with a custom converter that maps room codes to semantic names, reconstructs room polygons from boundary segments, and computes geometric attributes such as area, width, and height. The polygon vertices are expressed in absolute coordinates after scaling from pixels to meters. Each floor plan is converted into a set of spaces, each assigned a unique identifier and associated numeric attributes (e.g., area, width, height), together with the bubble diagram encoded as an adjacency list. We obtain each bubble diagram from the interior door connectivity graph produced by our \textit{RPLAN} conversion pipeline and represent it as an adjacency list stored in a JSON dictionary. Each key represents a room identifier, and each value is an array of room identifiers that are directly connected via interior doors. The front door is modeled as a special space that connects to exactly one room. Any sample whose adjacency list is disconnected or that contains rooms lacking valid polygons is filtered out.

Applying this pipeline yields four room-count specific datasets: 8-room with 53,001 training, 8,596 test, and 8,597 validation examples; 7-room with 44,859 training, 12,667 test, and 12,668 validation; 6-room with 47,955 training, 11,119 test, and 11,120 validation; and 5-room with 65,000 training, 2,597 test, and 2,597 validation.

The \textbf{RPLAN} dataset is released under a restricted-access, research-only data-use agreement. It is constructed from anonymized floor plans that remove user and privacy information, and the authors state that the underlying plans were curated to avoid copyright issues. Access is granted only via the authors’ official request process, and the terms explicitly limit use to non-commercial research and academic purposes and prohibit redistribution of the dataset in any way or format (in whole or in part). Accordingly, we do not redistribute RPLAN, nor any processed versions, subsets, annotations, or other derivative data products that contain any portion of the original dataset. Any work derived from RPLAN must follow the same conditions, including non-commercial academic use only and no redistribution; new users must obtain access directly through the authors’ official channel.

\subsection{Training and Inference Details}
\label{sec:training_details}

In the first training stage, we fine-tune the Llama-3.3-70B-Instruct \citep{grattafiori2024llama3} backbone using 4-bit quantization and adapter-based PEFT (LoRA) \citep{hu2021lora}. We configure LoRA with rank $r=64$, $\alpha = 128$, dropout $0.1$, and a learning rate of $1e-4$. Training is distributed across a 6-node Slurm cluster (each node: 4 × NVIDIA H100 80 GB). We pack the examples into a context window of 6k tokens, use a device batch size of 2, and train for two epochs. For the most complex 8-room floor plan generation task, this requires up to four hours, and the same amount of time or less on smaller-room tasks. 

We initialize the second stage from the supervised fine-tuning checkpoint, obtained by merging the SFT LoRA adapter into the base model. We then train with GRPO using TRL\footnote{\url{https://github.com/huggingface/trl}}. Training runs on a 7-node Slurm cluster with 4 NVIDIA H100 80\,GB GPUs per node: 6 nodes are used for distributed optimization, and 1 node hosts a dedicated vLLM server for rollout generation \citep{kwon2023efficient}. We use a per-device batch size of 1 and sample 4 generations per prompt. Optimization uses AdamW with learning rate $10^{-6}$, clipping parameter $0.2$, and KL coefficient 0.04. Rollouts are generated with temperature 0.9, top-$p$ 1.0, and maximum prompt and completion lengths of 4096 tokens. Validation is performed every 100 steps on 200 sampled validation examples, and checkpoints are selected based on validation reward. In practice, the best model is typically the first checkpoint evaluated at step 100, which on our hardware corresponds to at most about 2 hours of wall-clock time per task. This makes the GRPO stage shorter than supervised fine-tuning.

At inference time, unless otherwise stated, we use temperature 0.7, top-p 0.9, and a best-of-10 strategy. Candidates are ranked by minimum overlap area and, in case of ties, minimum Compatibility score.

Below is the exact prompt we use to condition the model:

\lstset{
  breaklines=true,
  breakatwhitespace=false,   
  columns=fullflexible,
  keepspaces=true,
  showstringspaces=false
}

\begin{lstlisting}[
  basicstyle=\scriptsize\ttfamily,
  caption={},
  label={lst:system_prompt}
]

<|begin_of_text|><|start_header_id|>system<|end_header_id|>You are a state-of-the-art floor-plan generator that translates JSON specifications and connectivity requirements defined by a bubble diagram into precise, optimized layouts. 
Your algorithm considers each room's dimensions, proportion, and desired adjacencies to produce an efficient arrangement that maximizes usable space while honoring all constraints.
Your top priority is that no two room polygons ever overlap. Rooms must be strictly disjoint, doors may touch room boundaries, but room interiors must never intersect.  
Your output must be a JSON object, where `output` key contains:
- `room_count`: the total number of room entries  
- `spaces`: a list of mixing rooms and doors. Each room or door entry must include:
   - `id`: formatted as `<room_type>|<unique_index>` (e.g. `"bedroom|2"` or `"interior_door|0"`)  
   - `room_type`: the room type (e.g. `"living_room"`, `"kitchen"`, etc.)
   - `area` in square meters (all positive numbers)  
   - `floor_polygon`: an ordered list of `{x: , y:}` vertices defining a simple polygon  
Additional rules:
- **Absolute non-overlap**: no two room polygons may share any interior point under any circumstances.
- Every adjacency in the bubble diagram must be bridged by exactly one door.  
- Every `id` used in the bubble diagram and on any door must appear in the `rooms` list.  
Return only a JSON object containing an `output` key without extra commentary or explanation.<|eot_id|>
\end{lstlisting}

\subsection{Compatibility, Realism and Diversity Metrics}
\label{sec:metrics_appendix}

\textbf{Compatibility} is computed as the graph edit distance \citep{sanfeliu1983distance} between the input bubble diagram and the bubble diagram reconstructed from the generated floor plan JSON. A lower score indicates higher consistency with the specified connectivity, with a score of 0 meaning a perfect match. In this sense, Compatibility can be interpreted as the number of connectivity mistakes in the generated floor plan, making it the most direct measure of whether the model satisfies the user-defined connectivity constraints.

\textbf{Realism} is measured through feedback from a set of volunteers: each person reviews 10 randomized pairs of layouts (one ground truth, one generated) and selects the more realistic layout, or indicates that both are equally realistic. For each pair, we assign  $+1$ if the volunteer selects the generated floor plan, $-1$ if they select the ground-truth floor plan, and $0$ if they judge them equally realistic. We report the mean of these scores across all comparisons and all volunteers. 
Values near zero indicate that, on average, generated layouts are indistinguishable from ground truth.

\textbf{Diversity} is measured using the Fréchet Inception Distance (FID) \citep{heusel2017gans}, which compares the feature distributions of generated floor plan images and ground truth floor plan images. Rather than relying on pixel-level differences, FID computes the mean and covariance of deep feature representations to assess the similarity between the two distributions. A lower FID means that the generated floor plans have feature statistics closer to those of the ground truth, indicating that the model captures both the diversity and the overall distribution of the reference data more effectively. We compute FID using a custom visualization pipeline that mimics the \textit{HouseDiffusion} visualizer on our data.

\subsection{Full Main Results with Standard Deviations}
\label{sec:full_gen_performance}
This section reproduces Table \ref{tab:gen_performance} and adds standard deviations. Table \ref{tab:compatibility_only} reports mean $\pm$ standard deviation for Compatibility on the five to eight-room tasks, using the same evaluation setup as in the main results. Table \ref{tab:realism_diversity_only} reports mean $\pm$ standard deviation for Realism and Diversity under the same setup; the only difference from the main results is the inclusion of dispersion values for completeness.

\begin{figure*}[t]
    \centering
    \includegraphics[width=0.95\linewidth]{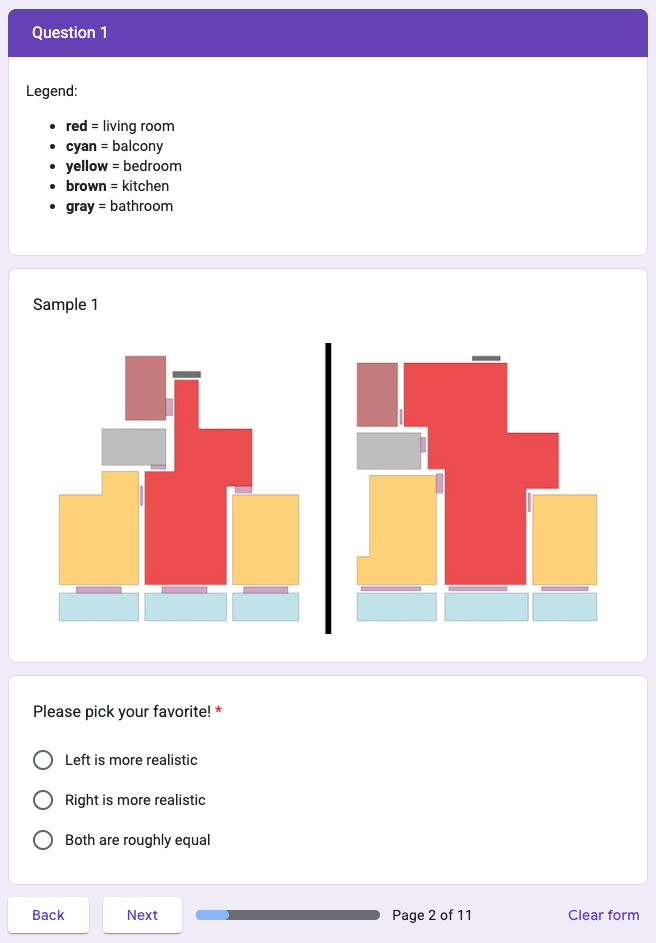}
    \caption{Example comparison from our Realism feedback exercise, showing a ground truth floor plan and a generated one from our method for the same bubble diagram. In this case, ground truth is left. The order was randomized across comparisons.}
    \label{fig:survey-screen}
\end{figure*}

\subsection{Realism Feedback Setup}
\label{sec:realism-survey}

Instructions provided to volunteers at the beginning of the Realism feedback exercise:

\begin{displayquote}\itshape

Please help us decide if the AI-generated floor plans look as realistic as the ground-truth ones.

What we want you to do is for each of the following 10 floor plans to decide if the one on the left or the one on the right looks more realistic, or they both look more or less the same. 

Notes: 
\begin{itemize}
    \item There are doors between rooms (marked in purple) and entrance doors (marked in dark gray) that are placed pretty arbitrarily, even in the GT dataset. 
    \item There are gaps between rooms (also known as "walls") of varying thickness, even within the same floor plan. This is normal.
    \item Floor plans only very rarely end up perfectly square, so don't look for that.
    \item Important: We are not asking you to detect which ones are AI-generated. We are asking you to pick which one you find more plausible or realistic.
    \item Important: Both being equal is a perfectly fine response.
    \item We will not elaborate what "realistic" or "plausible" means. Just use your best judgment.
\end{itemize}

\end{displayquote}

Figure \ref{fig:survey-screen} shows one example comparison from the feedback interface. We recruited postgraduate student volunteers through our academic network and did not provide monetary compensation. The feedback exercise was anonymous and did not collect personally identifiable information.

\subsection{Potential Risks and Mitigations}
Automating parts of the floor plan design process may contribute to job displacement or devaluation of professional expertise in architecture, drafting, real estate, and related services. As such systems become more capable, organizations may be incentivized to reduce human involvement, shifting labor demand away from early-stage drafting and toward fewer, higher-leverage roles. \textbf{Mitigation:} We position the system as an assistive tool intended to augment professionals rather than replace them. In any deployment, we recommend (i) treating outputs as drafts that require expert review and sign-off, (ii) providing training and upskilling resources to help practitioners integrate generative tools into established workflows, and (iii) designing interfaces that preserve human control (e.g., editable constraints, transparent validation reports, and explicit handoff points to licensed professionals).

Generated floor plans may fail to satisfy building codes, accessibility requirements, structural constraints, or domain-specific best practices, even when they satisfy the limited constraints we verify. If used without expert oversight, such outputs could contribute to unsafe designs, construction errors, and downstream legal liability, and could negatively affect occupants' quality of life. \textbf{Mitigation:} We recommend human-in-the-loop use with clear disclaimers that outputs are not certified designs. For higher-stakes settings, the pipeline should incorporate rigorous automated checks (e.g., egress and circulation, accessibility, minimum clearances, code rule sets), post-generation geometric repair, and professional review by qualified architects and engineers. 

\subsection{Behind the scenes}
In this section, we guide readers through the complete research process, including the experiments and approaches that did not work, sharing insights and lessons that may benefit other researchers.

\textbf{ProcTHOR.} We explored augmenting our training dataset with synthetic houses from ProcTHOR \citep{deitke2022procthor}, whose floor plan generator first samples an outer shell by iterative boundary cuts and then subdivides it into rooms using the recursive layout method of \citep{lopes2010constrained}. In that procedural generator, spaces are organized hierarchically into functional zones (public vs.\ private) before being split into individual rooms, which aligns with the typical day/night zoning principle. 

Despite these attractive properties, qualitative inspection revealed frequent topological pathologies that conflict with our target distribution: for example, bathrooms acting as corridors connecting a kitchen and a bedroom, or four-bedroom layouts where one must pass through multiple bedrooms to reach the last. Given the prevalence of such issues, we elected not to use ProcTHOR as a data source and relied instead on curated real-world layouts.

\textbf{Llama-3.1-8B-Instruct.} We first tried Llama-3.1-8B-Instruct as a backbone. Even with very large LoRA adapters (rank $r\in\{256,512\}$ and $\alpha\in\{128,256\}$), the model did not yield reliable results. At inference time, it often fell into a repetition loop that emitted the same value sequence for one polygon vertex array within a room, producing degenerate or invalid geometry. Because this failure mode persisted, we discarded the 8B variant for this application and adopted a Llama-3.3-70B-Instruct backbone instead.

\textbf{Few-shot Prompting Alone Is Not Enough.}
We evaluated few-shot prompting with state-of-the-art backbones, including GPT\mbox{-}4o, OpenAI o3, and QwQ\mbox{-}32B. None consistently produced a valid floor plan JSON. Few-shot prompting alone does not provide the built-in structure needed to enforce bubble diagram connectivity and numerical constraints jointly. Even with careful prompt engineering and schema exemplars, outputs remained brittle. Typical failures included non-closed polygons, self-intersections, duplicated or missing rooms, violations of bubble diagram connectivity, numerical drift in room areas, repetition loops that copied a single coordinate across a polygon, and schema hallucinations. 

We therefore adopted supervised fine-tuning, followed by reinforcement learning with verifiable rewards. That said, we do not rule out the possibility that future models with stronger structured generation capabilities may make few-shot prompting alone sufficient for this task.

\textbf{Reward Hacking.} We initially optimized only the Connectivity Reward. The model quickly learned to game this objective: it collapsed geometry into tiny rooms so that the reconstructed adjacency matched the prompt, yielding near-zero Compatibility while violating requested areas and hurting Realism. 


\end{document}